\def\year{2024}\relax
\documentclass[letterpaper]{article} 
\usepackage{aaai24}  
\usepackage{times}  
\usepackage{helvet}  
\usepackage{courier}  
\usepackage[hyphens]{url}  
\usepackage{graphicx} 
\usepackage{natbib}  
\usepackage{caption} 
\DeclareCaptionStyle{ruled}{labelfont=normalfont,labelsep=colon,strut=off} 
\usepackage{algorithm}
\usepackage{algorithmic}

\usepackage{newfloat}
\usepackage{listings}
\usepackage{xcolor}
\usepackage{multirow}

\floatstyle{ruled}
\newfloat{listing}{tb}{lst}{}
\floatname{listing}{Listing}
\title{CPL-NoViD: Context-Aware Prompt-based Learning for Norm Violation Detection in Online Communities}
\author{
    Zihao He, Jonathan May, Kristina Lerman
}
\affiliations{
USC Information Sciences Institute\\
zihaoh@usc.edu, \{jonmay, lerman\}@isi.edu

}

\usepackage{bibentry}

\begin{document}

\maketitle

\begin{abstract}
Detecting norm violations in online communities is critical to maintaining healthy and safe spaces for online discussions. Existing machine learning approaches often struggle to adapt to the diverse rules and interpretations across different communities due to the inherent challenges of fine-tuning models for such context-specific tasks. In this paper, we introduce \textbf{C}ontext-aware \textbf{P}rompt-based \textbf{L}earning for \textbf{No}rm \textbf{Vi}olation \textbf{D}etection (CPL-NoViD), a novel method that employs prompt-based learning to detect norm violations across various types of rules. CPL-NoViD outperforms the baseline by incorporating context through natural language prompts and demonstrates improved performance across different rule types. Significantly, it not only excels in cross-rule-type and cross-community norm violation detection but also exhibits adaptability in few-shot learning scenarios. Most notably, it establishes a new state-of-the-art in norm violation detection, surpassing existing benchmarks.  Our work highlights the potential of prompt-based learning for context-sensitive norm violation detection and paves the way for future research on more adaptable, context-aware models to better support online community moderators.
\end{abstract}

\section{Introduction}
Online communities have emerged as important platforms for social interaction, information sharing, and collaboration \cite{boyd2007social, wellman2018network, ellison2007benefits, he2024reading}. 
The ease of access to global online communities creates numerous advantages for society: it democratizes the generation and dissemination of information \cite{cropf2008benkler}, diminishes the influence of conventional gatekeepers in determining which information receives attention \cite{shirky2008here}, facilitates enhanced learning from diverse experiences \cite{iyer2009more}, and sparks large-scale protest movements \cite{tufekci2017twitter}. However, the very mechanisms that generate societal advantages also give rise to distinct vulnerabilities. Exchanging diverse perspectives within globally-connected online communities invites discord, which malicious actors and anti-social trolls exploit to disrupt conversations, spread misinformation, and increase polarization \cite{marwick2017media, del2016spreading}. 
Maintaining a safe and inclusive environment for all users in the rapidly growing online communities is necessary to maintain trust, promote democracy and safeguard mental health~\cite{lukianoff2019coddling}.

A popular approach to creating a safe and inclusive online environment involves detecting and moderating norm violations, which refers to any behavior that deviates from the community's established rules and expectations~\cite{danescu2013no, jhaver2019human, rajadesingan2020quick}. These norms can range from prohibiting hate speech and personal attacks to enforcing specific content formatting and sharing guidelines~\cite{chandrasekharan2017you, fiesler2018reddit}. Consequently, community moderators bear the responsibility of detecting and eliminating content that breaches the rules, irrespective of whether the violations are deliberate or arise from a limited understanding of the community's norms.

The sheer volume of content generated in online communities presents a significant challenge for moderators. For instance, Reddit saw more than 430 million posts and 2.5 billion comments posted in 2022 alone\footnote{https://www.demandsage.com/reddit-statistics/}. To cope with this deluge of content, moderators often turn to technological solutions such as such as machine learning algorithms and natural language processing tools, to highlight comments that might infringe upon community rules ~\cite{park2021detecting, jhaver2018online, davidson2017automated}. Although machine learning methods have been devised to identify inappropriate content, they mainly concentrate on automatically detecting the most egregious social violations, like toxicity and hate speech~\cite{Detoxify}. These approaches often fail to capture the diversity of norm violations, which go beyond hate speech to include a wide range of rule types. This diversity not only lies in the type of rules but also in the communities themselves. Each community has its own unique culture, norms, and interpretations of rules, making the development of universally applicable detection methods a challenging task. Yet, it is important to note that while a universal approach is often inadequate due to the unique nature of each community, a method that is adaptable and context-aware can be broadly applicable. 

\begin{figure*}[ht]
    \centering
    \includegraphics[width=0.98\textwidth]{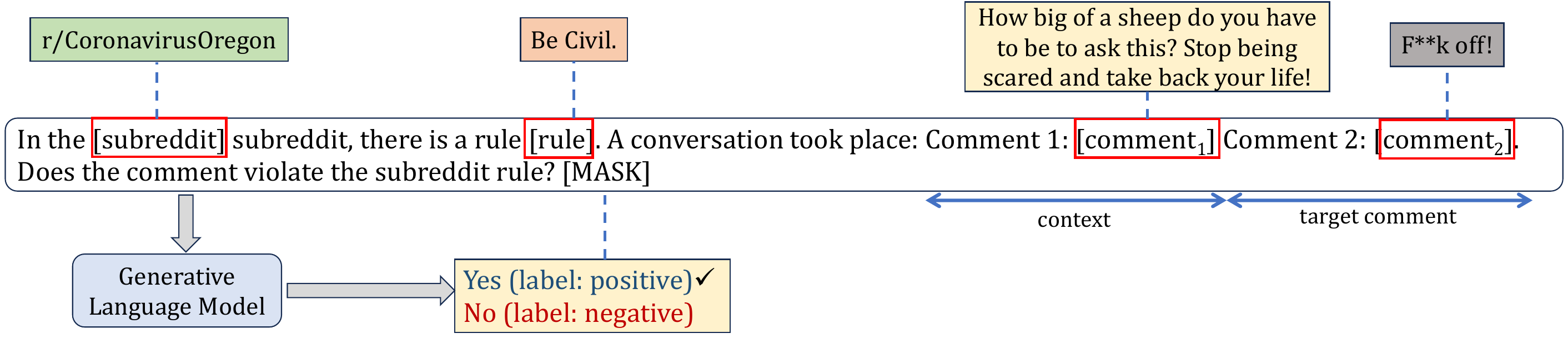}
    \caption{The architecture of CPL-NoViD. It is designed for prompt-based learning and integrates the conversation context into the prompt for a holistic understanding.}
    \label{fig:framework}
\end{figure*}

Given this background, our paper proposes a novel approach to norm violation detection that leverages prompt-based learning. We depart from the traditional binary classification problem formulation and instead frame the task as prompt-based learning. Specifically, for a target comment posted within an online discussion, we create a prompt in the format: ``In the [subreddit] subreddit, there is a rule: [rule]. A conversation took place: Comment 1: [comment$_1$] Comment 2: [comment$_2$] .... Comment n: [comment$_n$]. Does the last comment violate the subreddit rule? [MASK]''. The model is then expected to answer either ``Yes'' or ``No.'' Our approach, which we refer to as \textbf{C}ontext-aware \textbf{P}rompt-based \textbf{L}earning for \textbf{No}rm \textbf{Vi}olation \textbf{D}etection (CPL-NoViD), demonstrates better performance compared to baseline in detecting violations of rules of various types. Furthermore, our innovative application of prompt-based learning not only improves upon existing models but also establishes a new state-of-the-art in norm violation detection. The results affirm the efficacy of CPL-NoViD and its potential to significantly enhance content moderation across diverse online communities.

The primary motivation behind our CPL-NoViD is to address the limitations of existing methods in handling the diverse and context-sensitive nature of norm violations. The baseline method relies on a recurrent neural network (RNN) to incorporate context, which may suffer from gradient vanishing/exploding issues and introduces additional parameters to learn. In contrast, our approach seamlessly integrates context through natural language in the prompt, resulting in a more pronounced gain in context awareness. Moreover, our method exhibits significantly better generalizability in cross-rule-type and cross-community norm violation detection tasks, where the model is trained on data excluding the rule or community and tested on data solely from the rule or community. This adaptability is essential for building models that can cater to the unique norms and rules of various online communities. Finally, we explore the potential of our method in few-shot learning cases, where the model is exposed to only a limited number of training examples. In these scenarios, our method significantly outperforms the baseline, highlighting its utility in situations where obtaining a large amount of labeled data may be challenging or impractical.

In summary, our work aims to contribute to the ongoing efforts to develop advanced natural language processing techniques for online community moderation. By proposing CPL-NoViD, we hope to pave the way for more effective and adaptable approaches to norm violation detection that can accommodate the diverse and context-sensitive nature of conversations in online communities, establishing the new state-of-the-art. We believe that our work not only addresses the limitations of existing methods but also provides valuable insights and directions for future research in this area. As the digital landscape continues to evolve, it is increasingly important to develop robust and adaptable solutions that can support moderators and foster safe, inclusive, and thriving online communities. By focusing on context-awareness, generalizability, and the potential for few-shot learning, our work demonstrates the potential of prompt-based learning in tackling the complex challenges associated with norm violation detection and online community moderation.

\section{Related Work}

\paragraph{Online Community Moderation} 
Online community moderation is a well-studied field with a significant emphasis on identifying and mitigating various forms of norm violations. Tools like Automoderator~\cite{jhaver2019does} have been proposed, which use regex-based techniques to aid human moderators in managing large online communities. Additionally, efforts have been made to address specific types of harmful content such as toxicity and hate speech~\cite{wulczyn2017ex, davidson2017automated}. The Perspective API, for instance, provides an AI-based solution for toxicity detection, but its limited scope and the risk of overblocking make it less suitable for nuanced community moderation tasks~\cite{chandrasekharan2017you}. Acknowledging the contextual nature of community rules, context-aware models have been developed~\cite{yang2017identifying}. However, there is a clear need for more effective, adaptable, and nuanced solutions like our proposed CPL-NoViD, to address the unique norms and rules across diverse online communities.

\paragraph{Machine Learning for Norm Violation Detection} 
Machine learning has been extensively used for norm violation detection in online communities. Much of the existing research focuses on identifying content that violates community guidelines, such as hate speech, toxicity, and harassment \cite{davidson2017automated, park2018reducing, founta2018large, xu2012learning}. These approaches often employ traditional machine learning techniques, such as support vector machines (SVMs) and logistic regression, as well as deep learning methods like convolutional neural networks (CNNs) and recurrent neural networks (RNNs) \cite{nobata2016abusive, badjatiya2017deep, gamback2017using}. More recently, large-scale pre-trained language models, such as BERT and GPT, have shown remarkable performance in various natural language understanding tasks, including norm violation detection \cite{devlin2019bert, radford2018improving, wang2017liar}. However, most of these methods do not account for the diverse nature of community rules and their varying interpretations across different communities. As a result, they may struggle to adapt to the unique norms and guidelines of individual online communities, underscoring the need for more context-aware and community-specific approaches to norm violation detection. \citet{park2021detecting} create NORMVIO which is a norm violation detection dataset on Reddit, including additional context beyond the norm-violating comment itself; the authors also introduce a series of models for detecting norm violations, where context is encoded by RNN-based architecture that is likely to suffer from gradient vanishing/exploding issues and introduces additional parameters to learn.

\paragraph{Prompt-based Learning in NLP} 
Prompt-based learning has gained significant attention in the field of text classification as a promising approach for leveraging pre-trained language models in a controlled manner \cite{gao2021making, zhangdifferentiable, liu-etal-2022-p, zhang2022prompt}. The background for prompt-based learning stems from the success of transformer-based models like GPT-3 \cite{brown2020language}, which have shown remarkable performance on a wide range of natural language processing tasks. However, these models often require large amounts of labeled data for fine-tuning, which can be expensive and time-consuming to obtain. Prompt-based learning aims to address this limitation by providing a structured and interpretable way to guide the model's behavior using prompts or template-based inputs \cite{bao2020few}. By designing specific prompts, researchers can direct the model's attention towards relevant information and elicit desired responses. This approach not only reduces the reliance on extensive labeled data but also allows users to have more control and transparency over the model's decision-making process \cite{schick2021s}.
Our work leverages the advantages of prompt-based learning to create a flexible and adaptable method for detecting norm violations in online communities.

\section{Methodology}
In this section, we delve into the methodology used in this study, articulating the problem setup, the motivation behind the adoption of prompt-based learning, and our approach towards incorporating context into the model using natural language. Figure \ref{fig:framework} provides a schematic representation of the overall framework of our proposed method.

\subsection{Task Formulation}
Online communities, such as Reddit, function based on conversation threads, denoted by $c$. Each thread comprises $n$ comments $\{t_1, t_2, ...., t_n\}$, with the last comment $t_n$ serving as the target comment and the preceding $n-1$ comments $\{t_1, t_2, ...., t_{n-1}\}$ providing the contextual backdrop. Each subreddit, denoted by $s$, operates under a specific rule set, represented by the rule text $r$. 

The task at hand is to determine whether the target comment $t_n$ from a thread $c$ violates the rule text $r$ for a given subreddit $s$. It is crucial to highlight that this classification process is rule-specific; the comment is not evaluated for its general violation, but for its compliance with the specific subreddit's rule.

\subsection{Prompt-based Learning for Norm Violation Detection}
\label{sec:prompt-learning}
In an earlier work, \citet{park2021detecting} propose the usage of a fine-tuned pretrained language model to encode the comments. The model predicts the violation based on the encoded representation of the comments, rule text, and subreddit name. However, such a classification-based fine-tuning process can require a significant amount of labeled data and may not provide optimal domain-transferability \cite{schick2021s} due to the difference in the pretraining task (language modeling) and the fine-tuning task (classification).

To address these challenges, we introduce a novel approach that uses prompt-based learning. The idea revolves around designing prompts that seamlessly embed the subreddit name, rule, and comment into a natural language question. The prompt, formatted as ``In the [subreddit] subreddit, there is a rule: [rule]. A comment was posted: [comment]. Does the comment violate the subreddit rule? [MASK]'', sets the stage for the model to predict if a violation has occurred. We use ``Yes'' and ``No'' as the verbalizer for the positive and negative class respectively. By leveraging the inherent capabilities of language models in understanding language constructs, this approach aims to improve domain-transferability and reduce the requirement for extensive labeled data.

\subsection{Incorporating Context as Natural Language Prompts}
\label{sec:natural-lang-context}
Incorporating context into the model presents another challenge. \citet{park2021detecting} propose using LSTM to encode each preceding comment, attempting to capture the interdependencies between the context and the comment. However, this approach has two main drawbacks: 
1) LSTMs are prone to the vanishing gradient problem \cite{hochreiter1997long}, especially for long sequences, making it difficult to effectively capture long-range dependencies.
2) Tuning the hyperparameters of LSTMs, such as the number of layers and hidden units, is a laborious and expertise-driven task \cite{bergstra2012random}, adding to the complexity of the model.

Additionally, encoding the target comment along with each preceding comment leads to repetitive inclusion of the rule text and subreddit name, leading to inefficient encoding due to redundant meta-information.

To address these limitations, we propose an innovative methodology to incorporate context into the prompt using natural language. The context-aware section of the prompt is represented as ``A conversation took place: Comment 1: [comment$_1$] Comment 2: [comment$_2$] .... Comment n: [comment$_n$]''. By embedding the context directly within the prompt, we leverage the inherent capability of state-of-the-art language models to understand contextual relationships and dependencies. This enables more accurate and adaptable detection of norm violations. 
Furthermore, this approach eliminates the need for intricate hyperparameter tuning and manual experimentation, thereby streamlining the model training process and making it more efficient.

\subsection{Context-aware Prompt-based Learning for Norm Violation Detection}

Building upon the foundation of prompt-based learning and the strategy of natural language context incorporation, we introduce our proposed model - \textbf{C}ontext-aware \textbf{P}rompt-based \textbf{L}earning for \textbf{No}rm \textbf{Vi}olation \textbf{D}etection (CPL-NoViD). 
The final prompt used for finetuning the language model is structured as follows: ``In the [subreddit] subreddit, there is a rule: [rule]. A conversation took place: Comment 1: [comment$_1$] Comment 2: [comment$_2$] .... Comment n: [comment$_n$]. Does the last comment violate the subreddit rule? [MASK]''. 


For any given conversation thread $c$ consisting of $n$ comments, the subreddit of the thread $s$, and the rule text $r$, this prompt is populated using its respective comments. The objective is for the language model to ascertain whether a violation occurred and respond appropriately: "Yes" if a breach is detected, and "No" otherwise.
In detail, the [MASK] token is anticipated by the model to be any token within its vocabulary. We then extract the likelihoods for the predictions "Yes" and "No", optimizing these probabilities with a binary cross-entropy loss. This entire mechanism is visually represented in Figure \ref{fig:framework}.

This approach transforms the problem into a language modeling task, aligning it with the pretraining of the model. This alignment, in combination with the integration of the context into the prompt, enables CPL-NoViD to perform norm violation detection with improved accuracy and efficiency, without the need for extensive labeled data or complex hyperparameter tuning. 

The innovation within CPL-NoViD lies in its dual components: context learning and prompt-based learning. 
While CPL-NoViD effectively encodes context within natural language prompts, it is important to note that its successful application does not rely on the presence of context. This is because prompt-based learning, by its nature, offers great flexibility. As we will demonstrate in our experiments, CPL-NoViD maintains superior performance over baseline models even in the absence of context, relying solely on its prompt-based learning capability.

\section{Experiments}

\subsection{Data}
We experiment on NORMVIO \cite{park2021detecting},  a collection of 52K conversation threads on Reddit. This dataset includes 20K conversations whose last comment was removed by online moderators, along with  the text of the violated rule(s) that triggered the removal. These 20K conversations are used as positive (moderated) examples. To create negative examples, \citet{park2021detecting} came up with a control set comprising 32K paired unmoderated conversation threads. The data is structured such that each moderated conversation is matched with at most two unmoderated conversations from the same post. As a result, each data instance in the dataset consists of a conversation thread, the subreddit of the thread, the text of the rule(s), and a binary label indicating whether the last comment in the thread violates the rule.

\begin{figure}[t]
    \centering
    \includegraphics[width=0.46\textwidth]{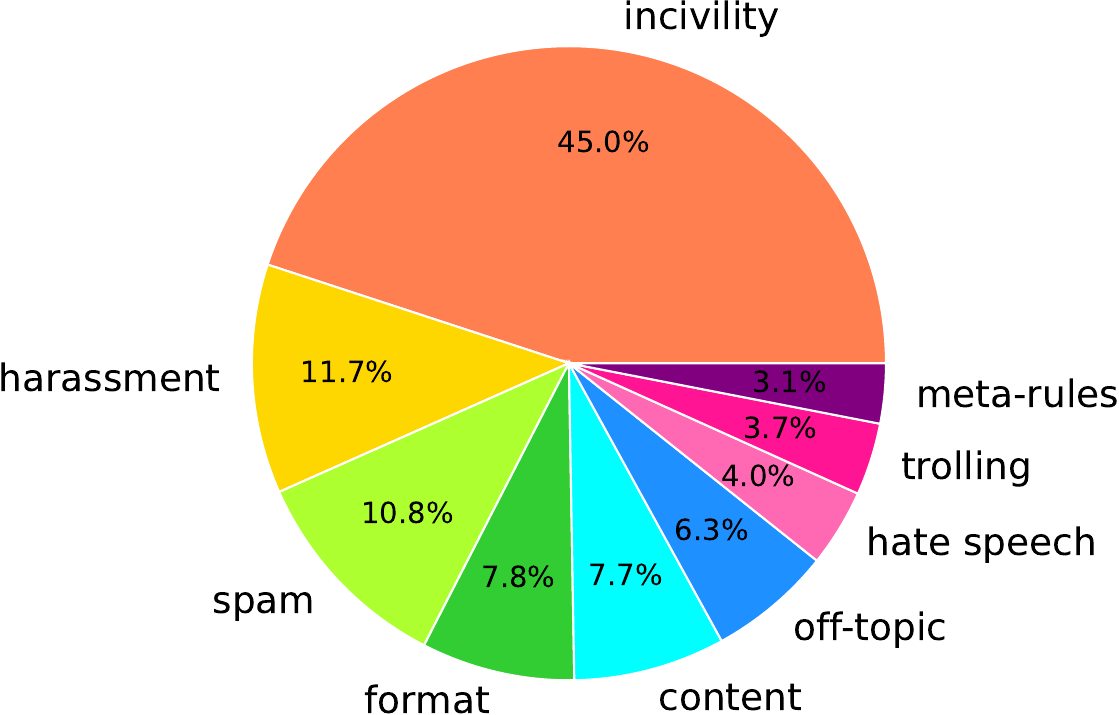}
    \caption{Sector diagram of different rule types in NORMVIO. The diagram is adapted from \citet{park2021detecting}.}
    \label{fig:rule_diag}
\end{figure}

NORMVIO divides rules into 9 coarse-grained rule types: incivility, harassment, spam, format, content, off-topic, hate speech, trolling, and meta-rules. For each conversation, the specified rule text could potentially fall under multiple rule types. The distribution of different rule types are shown in Figure \ref{fig:rule_diag}, where nearly half of the rules belong to incivility.

\begin{figure}[t]
    \centering
    \includegraphics[width=0.46\textwidth]{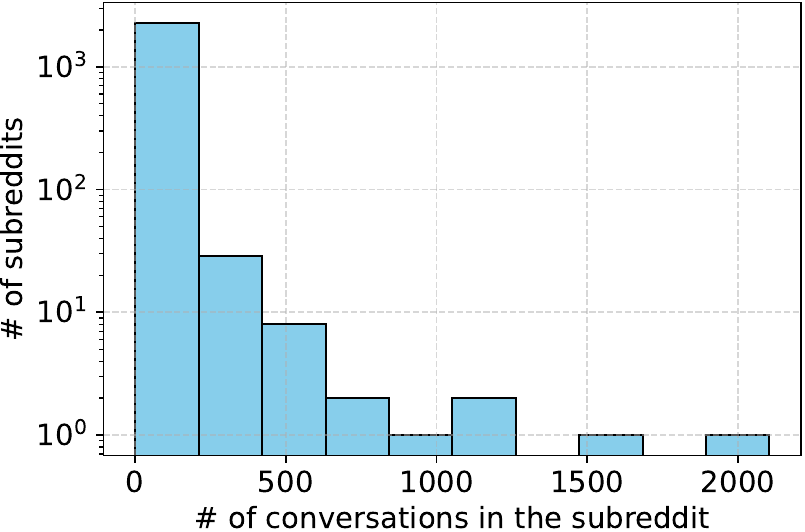}
    \caption{Distribution of number of conversations in each subreddit. The distribution is heavily long-tailed.}
    \label{fig:hist_n_convs}
\end{figure}

The dataset spans conversations from a diverse array of 2,310 unique subreddits. The distribution of conversation counts within each subreddit is visually depicted in Figure \ref{fig:hist_n_convs}. Upon examination, it is evident that the distribution exhibits a pronounced long-tail characteristic. This indicates that a majority of the subreddits are underrepresented in terms of training examples, potentially limiting the robustness of models trained solely on this dataset.
This characteristic of our dataset brings into sharp focus the issue of model generalizability. Given the dynamic nature of online communities, new subreddits are constantly emerging, and each may have its own unique set of norms and rules. A model's ability to adapt and perform well across a range of different subreddits, including those it was not specifically trained on, is therefore of paramount importance.
In the ensuing experimental section, we delve deeper into this issue, evaluating our proposed methodology's ability to generalize across diverse subreddits and discussing potential strategies for enhancing model generalizability in the context of norm violation detection.

\subsection{Comparative Baselines}
In order to evaluate our proposed model, we set a comparison against the following baselines:

\begin{itemize}
    \item \textbf{Majority}: This is a simple majority classifier that always predicts the most common class in the training data. It serves as a basic baseline for comparison.

    \item \textbf{Perspective}: This model uses the Perspective API\footnote{\url{https://perspectiveapi.com}}, to assign a toxicity score to each comment, independent of its contextual information. By adjusting the threshold value for each type of rule, the model optimizes for the highest F1 score on the development set.

    \item \textbf{BERT-LSTM} \cite{park2021detecting}: This model employs a pretrained BERT \cite{devlin2019bert} model to encode all the comments within a conversation. Following this, an LSTM context encoder is applied to capture the dependencies between the comments. A fully-connected layer is added on top of the hidden state from the last LSTM cell to generate the final prediction. The input for the BERT model includes comments, rule texts, and subreddit names, formatted as ``subreddit: [subreddit]. rule text: [rule text]. comment: [comment]. " 

    \item \textbf{T5-LSTM}: This model follows the same structure as BERT-LSTM, but substitutes BERT with T5 \cite{raffel2020exploring}, a generative language model, as the utterance encoder. Given that T5 has an encoder-decoder architecture, we solely use the encoder portion for this model.
\end{itemize}

We compare to two transformer-based models, BERT-LSTM and T5-LSTM, to provide a robust comparison with our proposed CPL-NoViD. BERT-LSTM is chosen given that BERT has been widely utilized and proven effective in classification tasks, making it a standard benchmark for comparison. It is also the existing state of the art method in norm violation detection on NORMVIO dataset \cite{park2021detecting}.

However, as our proposed CPL-NoViD is based on T5, we deem it necessary to also include T5-LSTM as a baseline to ensure a fair comparison. \textbf{While it is true that T5 is traditionally used for generation tasks and we understand that some may argue that using T5 in this way is not fully exploiting its potential, we will illustrate in the following subsections that even using it in a classification context by leveraging only its encoder portion, it outperforms BERT which is more suitable for the classification task.} 
We believe T5-LSTM offers a more accurate baseline comparison for CPL-NoViD, given that both methods use the same underlying model. Additionally, comparing CPL-NoViD with T5-LSTM helps to highlight the specific contributions and benefits of the prompt-based learning and context incorporation techniques that we introduce in CPL-NoViD.


\subsection{Experimental Setup and Implementation Details}
\label{sec:exp_setup}

Our CPL-NoViD model leverages T5 as the core generative language model. The implementation of all models, including the baselines and ours, is performed in PyTorch \cite{paszke2019pytorch}. The pretrained BERT and T5 models, specifically \emph{bert-base} and \emph{t5-base}, are sourced from the Huggingface library \cite{wolf2019huggingface}. The prompt-based learning framework is implemented using OpenPrompt \cite{ding2022openprompt}.

We limit the maximum conversation length to six comments, as it covers over 90\% of the dataset.\footnote{Empirical observations have shown that permitting longer conversations does not lead to improved model performance, and may even hinder the efficiency of training} In the case of conversation threads exceeding this length, we truncate them, preserving only the \emph{most recent} six comments. We follow the same 80-10-10 train/val/test random split of NORMVIO as in \citet{park2021detecting}.

The maximum utterance length is set differently for BERT-LSTM and T5-LSTM, and CPL-NoViD. For the former two, it is limited to 128 tokens, while for our CPL-NoViD model, it is extended to 384 tokens, considering the incorporation of context into a single utterance\footnote{About 85\% of comments are under 55 tokens in length. Given up to six comments per thread, we anticipated 330 tokens for comments alone. With additional tokens for prompt context, we approach roughly 360 tokens. Thus, 384 tokens were chosen to accommodate the majority of the dataset's conversations.}.

We adopt the Adam optimizer for model finetuning, setting the learning rate to $1e-4$. As for batch sizes, they vary across models to optimize GPU use: BERT-LSTM, T5-LSTM, and CPL-NoViD use batch sizes of 224, 200, and 44, respectively.

Training for all models spans 50 epochs and is conducted on a Tesla V100 GPU with 32GB memory. Model checkpoints that exhibit the best performance on the validation set are chosen for testing. To ensure robustness and reduce the impact of random initialization, we train each model with three different random seeds and report the average performance.

\begin{table*}[ht]
\def\arraystretch{1.2}
\addtolength{\tabcolsep}{-4.2pt}
\centering
\begin{tabular}{lcccccccccc}
\hline
\multicolumn{1}{c}{\textbf{Method}} & \textbf{macro} & \textbf{incivility} & \textbf{harassment} & \textbf{spam} & \textbf{format} & \textbf{content} & \textbf{off-topic} & \textbf{hate speech} & \textbf{trolling} & \textbf{meta-rules} \\ \hline
\textbf{Majority}                   & 62.6                             & 76.5                & 57.7                & 77.8          & 62.8            & 66.5             & 59.4               & 54.4                 & 55.7              & 52.8                \\
\textbf{Perspective}                & 52.6                             & 68.0                & 57.1                & 47.5          & 49.1            & 51.0             & 48.9               & 51.3                 & 49.0              & 51.2                \\
\textbf{BERT-LSTM}                  & 76.5$\pm$1.2                         & 79.7$\pm$0.9            & 78.5$\pm$0.8            & 79.6$\pm$0.6      & 75.3$\pm$1.0        & 72.7$\pm$4.4         & 75.1$\pm$2.7           & \textbf{80.4}$\pm$1.5             & 70.9$\pm$0.8          & 76.8$\pm$1.4            \\
\textbf{T5-LSTM}                    & 77.8$\pm$1.0                         & \textbf{81.5}$\pm$0.8            & 78.7$\pm$0.8            & 81.8$\pm$1.0      & \textbf{79.3}$\pm$0.8        & \textbf{74.8}$\pm$3.2         & 76.4$\pm$0.9           & 79.6$\pm$0.9             & 71.5$\pm$3.6          & 76.8$\pm$1.7            \\
\textbf{CPL-NoViD}                  & \textbf{78.2}$\pm$0.4                         & 80.4$\pm$0.4            & \textbf{80.4}$\pm$0.2            & \textbf{82.2}$\pm$0.4      & 77.3$\pm$1.8        & 72.2$\pm$2.5         & \textbf{78.6}$\pm$0.1           & 78.6$\pm$0.3             & \textbf{77.5}$\pm$1.2          & \textbf{78.9}$\pm$1.2            \\ \hline
\end{tabular}
\addtolength{\tabcolsep}{4.2pt}
\caption{F1 scores (\%) of norm violation detection on different rule types as well as the macro-averaged score. The top-performing scores in each category are highlighted in bold. While BERT-LSTM represents the existing state-of-the-art in norm violation detection, our proposed CPL-NoViD model surpasses its performance, thereby establishing a new benchmark in the field.}
\label{tab:results_overall}
\end{table*}

\subsection{Results and Analysis}

We present the norm violation detection results in Table \ref{tab:results_overall}. In terms of macro-F1 scores, our CPL-NoViD model surpasses all the baselines, establishing a new state-of-the-art performance in this task. Interestingly, the T5-LSTM model, which employs solely the encoder component of the T5 model, surpasses the performance of the BERT-LSTM model. This result reflects the inherent robustness of the T5 model in interpreting and processing textual data, even when it is deployed in a non-generative task setting.

An important observation is that CPL-NoViD exhibits lower fluctuations in performance, as shown by smaller standard deviations. We attribute this to the fact that prompt-based learning relies on the pretrained language model without introducing any new parameters, whereas BERT-LSTM and T5-LSTM require training an LSTM context encoder from scratch. Consequently, different parameter initializations can lead to larger performance fluctuations in the latter two models.

Our CPL-NoViD model demonstrates more pronounced performance gains on the ``trolling'' and ``meta-rules'' categories (6.0\% and 2.1\% improvement against the T5-LSTM baseline, respectively) compared to other rule types. These two categories constitute the smallest proportions among the nine rule types. As such, CPL-NoViD's enhanced ability to detect norm violations in these less-popular categories showcases its robust learning capabilities, even in cases with limited data availability.

While the ``hate speech'' category has been the primary focus of many NLP research efforts on abusive language detection \cite{jurgens2019just, zhang2018conversations, danescu2013no, almerekhi2020investigating}, it remains one of the least dominant rule types. Nevertheless, all three language model-based methods achieve commendable performance on this category. This can be attributed to the fact that hate speech is typically more easily identifiable, as it often involves explicit or implicit expressions of hostility, prejudice, or discrimination. 
Notably, BERT-LSTM slightly edges out other models in detecting hate speech''. This advantage may be attributed to BERT's architecture and training methodology, which could make it more attuned to the specific linguistic constructs found in hate speech, rather than its fundamental capability to capture semantic and syntactic patterns, which all models tested here possess.
For more nuanced rule types, such as ``spam'' and ``off-topic'', accurate detection demands a deeper understanding of context and domain knowledge. In these instances, our proposed CPL-NoViD method excels by leveraging context-aware prompts that provide natural language context. By effectively incorporating conversation history, CPL-NoViD is better equipped to discern subtleties and relationships between comments, which is crucial for detecting rule violations in these more complex categories.

Given the architectural similarities between BERT-LSTM and T5-LSTM, as well as the superior overall performance of T5-LSTM, we opt to use T5-LSTM as the sole baseline in subsequent subsections.

\subsection{Assessing Out-of-Domain Generalizability for Norm Violation Detection}

With the continued, rapid expansion of Reddit's user base and the concomitant rise in diverse subreddit communities, there is a pressing need for online moderation tools to effectively adapt to new, evolving rule types and newly minted subreddits. This necessitates a comprehensive assessment of our proposed model's out-of-domain generalizability. In this subsection, we undertake a series of experiments to evaluate the performance of our CPL-NoViD model and the baseline against unseen rule types and subreddits.

\begin{table*}[ht]
\def\arraystretch{1.2}
\addtolength{\tabcolsep}{-2.0pt}
\centering
\begin{tabular}{lccccccccc}
\hline
\multicolumn{1}{c}{\textbf{Method}} & \textbf{incivility} & \textbf{harassment} & \textbf{spam} & \textbf{format} & \textbf{content} & \textbf{off-topic} & \textbf{hate speech} & \textbf{trolling} & \textbf{meta-rules} \\ \hline
\textbf{T5-LSTM}  & 75.4$\pm$1.2            & 74.5$\pm$0.2            & 70.6$\pm$2.4      & 76.9$\pm$2.7        & \textbf{65.3}$\pm$1.4         & 68.6$\pm$0.8  & 77.3$\pm$0.8  & 67.5$\pm$1.3   & \textbf{78.8}$\pm$3.5   \\
\textbf{CPL-NoViD} & \textbf{76.0}$\pm$0.2            & \textbf{77.3}$\pm$1.7           & \textbf{74.2}$\pm$1.1      & \textbf{77.5}$\pm$0.6        & 62.4$\pm$4.2         & \textbf{72.4}$\pm$2.1           & \textbf{77.8}$\pm$1.8            & \textbf{74.0}$\pm$4.4          & 78.0$\pm$0.4           \\ \hline
\end{tabular}
\addtolength{\tabcolsep}{2.0pt}
\caption{F1 scores (\%) of norm violation detection on cross-rule-type norm violation detection. For each rule type, the model is finetuned on data excluding the rule type, and tested on data soley from it. The top-performing scores in each category are highlighted in bold. CPL-NoViD outperforms T5-LSTM in cross-rule-type generalizablity across most rule types.}
\label{tab:results_cross_rule}
\end{table*}

\begin{table*}[ht]
\def\arraystretch{1.2}
\centering
\begin{tabular}{lcccc}
\hline
\multicolumn{1}{c}{\textbf{Method}} & \textbf{r/CanadaPolitics} & \textbf{r/LabourUK} & \textbf{r/classicwow} & \textbf{r/Games} \\ \hline			
\textbf{T5-LSTM}                    & 71.4$\pm$7.2                  & 59.1$\pm$4.5           & \textbf{85.3}$\pm$1.7                    & 70.1$\pm$2.0         \\			
\textbf{CPL-NoViD}                  & \textbf{73.9}$\pm$2.3                 & \textbf{65.7}$\pm$6.7          & 82.6$\pm$2.4           & \textbf{79.1}$\pm$3.4         \\ \hline
\end{tabular}
\caption{Micro F1 scores (\%) of cross-community norm violation detection. For each community, the model is finetuned on data excluding the community, and tested on data only from it. The top-performing scores in each category are highlighted in bold. CPL-NoViD outperforms T5-LSTM in cross-community generalizablity across most communities.}
\label{tab:results_cross_comm}
\end{table*}

\subsubsection{Cross-Rule-Type Generalizability}
In this experiment, we test the model's performance on each of the nine rule types, removing the data of each rule type from the training set in turn. Each sample in the dataset consists of the conversation thread, the rule text, and the subreddit. The rule text is divided into one of the nine categories. 
To illustrate, when testing for ``harassment'', the training set is comprised of data from the other eight rule types, and the test set exclusively contains ``harassment'' data. The results of these cross-rule-type tests are detailed in Table \ref{tab:results_cross_rule}.

These results reveal that CPL-NoViD consistently outperforms or performs comparably to the baseline model in the majority of rule types, indicating superior cross-rule-type generalizability. The ability of CPL-NoViD to use the pretrained knowledge and context-aware natural language prompts effectively likely contributes to its strong performance. However, it is important to note that the performance varies across different rule types for both models, indicating the complexity of the task and the diversity of the rules. 

The evaluation results reveal that CPL-NoViD does not outperform the baseline model in the ``content'' and ``meta-rules'' categories. Upon examining these categories, we can infer potential reasons for this outcome.
In the ``content'' rule type, which includes low-quality content, NSFW, and spoilers, the rules are often intricately linked with subjective judgement and the specific standards of individual subreddit communities. For example, a rule such as ``no low-quality posts'' is inherently subjective and what constitutes as ``low-quality'' can vary significantly from one subreddit to another. Identifying violations of these rules may require a nuanced understanding of the specific quality standards and cultural norms of individual subreddits, which may not be fully captured by the generalized prompts used in CPL-NoViD. In such cases, the baseline model, which incorporates LSTM-based context encoding, could potentially have an advantage by being able to better model the unique characteristics of each subreddit.
``Meta rules'', on the other hand, govern aspects such as voting, moderation, enforcement, and adherence to ``reddiquette''. Violations of these rules are often more subtle and can involve indirect references or behaviors, such as downvoting. An example rule like ``no downvoting'' may be particularly challenging to identify as it requires understanding not just the text of a comment, but the actions and behaviors associated with it. This kind of meta-level understanding may not be fully captured by CPL-NoViD's text-based prompts and pretrained knowledge.

\subsubsection{Cross-Community Generalizability}

Given the sheer number of subreddits (2,310) included in the NORMVIO dataset, testing our model's performance on each is impractical. Therefore, to obtain a representative snapshot of its performance, we select four large subreddits from the top ten for this experiment: r/CanadaPolitics (2.1\%), r/LabourUK (1.2\%), r/classicwow (2.1\%), and r/Games (1.6\%). These subreddits span a wide range of topics, from politics and labor issues to gaming and general discussions. Given the highly skewed distribution of different rule types in the test data for each subreddit, wherein certain rule types are represented by only a few examples, we have chosen to use micro-F1 as our evaluation metric. The micro-F1 score provides a more reliable and stable measure of performance in such scenarios, as it aggregates the contributions of all classes to compute the average, and is thus not unduly influenced by the performance on the underrepresented classes. 
The results of these cross-community experiments are illustrated in Table \ref{tab:results_cross_comm}.

For the four evaluated communities, CPL-NoViD outperforms T5-LSTM in three out of four scenarios, namely r/CanadaPolitics, r/LabourUK, and r/games. The only exception is the r/classicwow subreddit, where T5-LSTM slightly outperforms CPL-NoViD. A potential reason for CPL-NoViD's superior performance in r/CanadaPolitics, r/LabourUK, and r/games could be the complexity of the discussion topics in these subreddits. These communities cover a wide range of subjects, including politics, labor issues, and various aspects of gaming, which require a nuanced understanding of context and intricate relationships between comments for effective moderation. CPL-NoViD, with its context-aware natural language prompts, might be better equipped to handle this complexity, leading to its superior performance.
On the other hand, r/classicwow is primarily focused on a single topic: the classic version of the game World of Warcraft.  CPL-NoViD's strength lies in leveraging context-aware prompts to handle complex and varied discussions. In a more uniform discussion environment like r/classicwow, this advantage might not be fully realized, leading to T5-LSTM's slight performance edge.

\subsubsection{Conclusion and Implications}

Our out-of-domain experiments underline CPL-NoViD's superior generalizability in both cross-rule-type and cross-subreddit settings. The model's prompt-based learning approach, combined with its adept incorporation of natural language context, enables it to effectively use pretrained knowledge, while also accurately capturing the complexities and dependencies within conversations.

Moreover, these results underscore CPL-NoViD's potential for real-world applications. By demonstrating its adaptability to the dynamic nature of online communities, where new rules and subreddits are constantly emerging, CPL-NoViD proves itself a promising solution for norm violation detection in online moderation, particularly for rapidly evolving platforms like Reddit.

\begin{table*}[ht]
\def\arraystretch{1.2}
\addtolength{\tabcolsep}{-3.5pt}
\centering
\begin{small}
\begin{tabular}{lcccccccccc}
\hline
\textbf{Method}    & \textbf{macro} & \textbf{incivility} & \textbf{harassment} & \textbf{spam} & \textbf{format} & \textbf{content} & \textbf{off-topic} & \textbf{hate speech} & \textbf{trolling} & \textbf{meta-rules} \\ \hline
\multicolumn{11}{c}{\textbf{n=10}} \\

\textbf{T5-LSTM}   & 42.0$\pm$18.2      & 43.3$\pm$13.9            & 39.8$\pm$15.4            & 41.2$\pm$27.3      & 44.5$\pm$20.8        & 45.0$\pm$17.4         & 40.2$\pm$25.9           & 41.3$\pm$13.4             & 36.4$\pm$21.7          & 46.3$\pm$10.4            \\
\textbf{CPL-NoViD} & \textbf{53.9}$\pm$3.0       & \textbf{52.7}$\pm$2.4            & \textbf{51.5}$\pm$5.9            & \textbf{56.6}$\pm$5.7     & \textbf{52.9}$\pm$5.8        & \textbf{52.5}$\pm$3.9         & \textbf{57.7}$\pm$2.8           & \textbf{54.4}$\pm$0.7             & \textbf{53.2}$\pm$3.1          & \textbf{53.8}$\pm$1.5            \\ \hline
\multicolumn{11}{c}{\textbf{n=50}}                                                                                                                                                                                         \\
\textbf{T5-LSTM}   & 44.7$\pm$20.1       & 46.6$\pm$16.0            & 44.6$\pm$18.1            & 41.5$\pm$27.6      & 46.5$\pm$22.5        & 46.8$\pm$18.9         & 41.3$\pm$26.8           & 46.4$\pm$15.6             & 41.4$\pm$24.3          & 48.1$\pm$11.7            \\									
\textbf{CPL-NoViD} & \textbf{59.9}$\pm$2.9       & \textbf{61.6}$\pm$0.5            & \textbf{62.0}$\pm$2.8            & \textbf{60.2}$\pm$5.2     & \textbf{56.2}$\pm$6.3        & \textbf{57.3}$\pm$1.9         & \textbf{61.4}$\pm$4.2           & \textbf{58.9}$\pm$2.3             & \textbf{57.2}$\pm$3.2          & \textbf{64.5}$\pm$4.7            \\ \hline
\multicolumn{11}{c}{\textbf{n=100}}                                                                                                                                                                                        \\
\textbf{T5-LSTM}                    & 77.8$\pm$1.0                         & \textbf{81.5}$\pm$0.8            & 78.7$\pm$0.8            & 81.8$\pm$1.0      & \textbf{79.3}$\pm$0.8        & \textbf{74.8}$\pm$3.2         & 76.4$\pm$0.9           & \textbf{79.6}$\pm$0.9             & 71.5$\pm$3.6          & 76.8$\pm$1.7            \\
\textbf{CPL-NoViD}                  & \textbf{78.2}$\pm$0.4                         & 80.4$\pm$0.4            & \textbf{80.4}$\pm$0.2            & \textbf{82.2}$\pm$0.4      & 77.3$\pm$1.8        & 72.2$\pm$2.5         & \textbf{78.6}$\pm$0.1           & 78.6$\pm$0.3             & \textbf{77.5}$\pm$1.2          & \textbf{78.9}$\pm$1.2            \\ \hline
\end{tabular}
\end{small}
\addtolength{\tabcolsep}{3.5pt}
\caption{F1 scores (\%) of few-shot learning performance of T5-LSTM and CPL-NoViD on norm violation detection, evaluated at different numbers of training instances per rule type. The best results are highlighted in bold. In terms of Macro F1-scores, CPL-NoViD outperforms T5-LSTM across different number of training instances.}
\label{tab:results_few_shot}
\end{table*}

\subsection{Few-Shot Learning for Norm Violation Detection}
The challenge of limited availability of labeled data for specific rule types or subreddits, especially in the context of newly instituted rules or emerging communities, is a common reality in real-world online platforms. Few-shot learning, which entails learning from a minuscule number of labeled examples, aims to tackle this very obstacle. In this subsection, we delve into a comparison between the few-shot norm violation detection capabilities of the baseline and CPL-NoViD.

To critically evaluate the models within the context of few-shot learning, we simulate a scarcity of labeled data through subsampling the training set for each rule type. To this effect, we generate three distinct training sets, each containing 10, 50, and 100 labeled examples per rule type respectively. The original test set is kept unchanged. Fine-tuning the models on these reduced training sets, we assess their performance on the unchanged test set. The outcomes of these assessments are recorded in Table \ref{tab:results_few_shot}.

As discerned from the results, CPL-NoViD maintains a consistent edge over the baseline model across all the few-shot learning environments, suggesting that CPL-NoViD is more adept at learning effectively from a smaller sample of examples. This can be ascribed to the combination of prompt-based learning and natural language context incorporation, which allow the model to more effectively use the existing knowledge embedded in the pre-trained language model, thereby requiring fewer labeled examples to achieve satisfactory performance.

Moreover, we notice a high variance in the performance of the baseline model, because when initialized with one of the random seeds the model fails to converge, resulting in an extremely low performance. This indicates a potential instability in the learning process when faced with fewer training examples. Interestingly, the performance of the baseline model exhibits negligible improvement when the number of training examples is increased from 10 to 50. In contrast, the performance improvement of CPL-NoViD is much more substantial, indicating that CPL-NoViD is better equipped to leverage additional training data to enhance its performance. This comparison further underscores the adaptability and robustness of CPL-NoViD in the face of limited data, a characteristic that is essential for its application in real-world, ever-evolving online platforms.

\begin{table*}[ht]
\def\arraystretch{1.2}
\addtolength{\tabcolsep}{-3.5pt}
\centering
\small
\begin{tabular}{lcccccccccc}
\hline
\textbf{Method}    & \textbf{macro}    & \textbf{incivility} & \textbf{harassment} & \textbf{spam}     & \textbf{format}   & \textbf{content}  & \textbf{off-topic} & \textbf{hate speech} & \textbf{trolling} & \textbf{meta rules} \\ \hline
\multicolumn{11}{c}{\textbf{max context size = 0 (no context)}}                                                                                                                                                                                   \\

\textbf{T5}   & 75.7$\pm$1.4          & 79.1$\pm$0.7            & 79.7$\pm$0.6            & 78.6$\pm$0.4          & 76.2$\pm$0.8          & \textbf{70.0$\pm$2.6} & 72.8$\pm$2.6           & \textbf{79.4}$\pm$2.6    & 70.3$\pm$2.1          & \textbf{74.8}$\pm$3.2   \\
\textbf{CPL-NoViD} & \textbf{75.9}$\pm$0.4 & \textbf{80.6}$\pm$0.1   & \textbf{80.4}$\pm$1.5   & \textbf{79.0}$\pm$0.4 & \textbf{77.7}$\pm$1.0 & 69.8$\pm$1.6          & \textbf{73.1}$\pm$0.9  & 75.7$\pm$2.1             & \textbf{72.5}$\pm$2.2 & 74.5$\pm$0.4            \\ \hline
\multicolumn{11}{c}{\textbf{max context size = 1}}                                                                                                                                                                                   \\ 
\textbf{T5-LSTM}   & 75.5$\pm$0.5          & 80.3$\pm$0.7            & 78.0$\pm$1.0            & 79.2$\pm$2.5          & 76.4$\pm$0.9          & 71.2$\pm$0.8          & 75.4$\pm$0.7           & 77.8$\pm$0.9             & 66.8$\pm$1.2          & 74.6$\pm$1.5            \\
\textbf{CPL-NoViD} & \textbf{77.5}$\pm$0.8 & \textbf{80.5}$\pm$0.2   & \textbf{80.1}$\pm$1.0   & \textbf{80.0}$\pm$1.0 & \textbf{76.5}$\pm$1.7 & \textbf{71.3}$\pm$2.0 & \textbf{76.9}$\pm$1.1  & \textbf{78.3}$\pm$0.8    & \textbf{74.2}$\pm$2.2 & \textbf{79.5}$\pm$1.3   \\ \hline
\multicolumn{11}{c}{\textbf{max context size = 2}}                                                                                                                                                                                   \\
\textbf{T5-LSTM}   & 77.1$\pm$0.2          & 80.2$\pm$0.2            & \textbf{79.1}$\pm$1.1   & 81.2$\pm$0.7          & 76.6$\pm$1.9          & \textbf{73.9}$\pm$1.8 & 76.3$\pm$1.7           & \textbf{80.4}$\pm$0.5    & 70.6$\pm$4.1          & 76.2$\pm$3.9            \\
\textbf{CPL-NoViD} & \textbf{77.9}$\pm$0.6 & \textbf{81.0}$\pm$0.7   & 79.0$\pm$0.1   & \textbf{81.8}$\pm$1.1 & \textbf{78.0}$\pm$2.1 & 72.3$\pm$0.3          & \textbf{77.5}$\pm$0.4  & 79.6$\pm$2.4             & \textbf{72.8}$\pm$0.4 & \textbf{78.9}$\pm$0.4   \\ \hline
\multicolumn{11}{c}{\textbf{max context size = 3}}                                                                                                                                                                                   \\

\textbf{T5-LSTM}   & 76.0$\pm$0.7          & 80.1$\pm$0.5            & 78.2$\pm$0.1            & 80.3$\pm$1.4          & 77.1$\pm$1.4          & 71.1$\pm$3.3          & 74.8$\pm$1.7           & \textbf{77.8}$\pm$2.6    & 68.7$\pm$1.3          & 76.1$\pm$2.9            \\
\textbf{CPL-NoViD} & \textbf{78.0}$\pm$0.3 & \textbf{81.0}$\pm$0.4   & \textbf{80.7}$\pm$1.6   & \textbf{82.0}$\pm$0.4 & \textbf{78.6}$\pm$1.3 & \textbf{72.5}$\pm$1.7 & \textbf{77.1}$\pm$2.3  & 76.6$\pm$0.5             & \textbf{75.1}$\pm$3.2 & \textbf{78.9}$\pm$3.0   \\ \hline

\multicolumn{11}{c}{\textbf{max context size = 4}}                                                                                                                                                                                   \\

\textbf{T5-LSTM}   & 77.6$\pm$0.8          & 80.2$\pm$0.3            & 79.4$\pm$0.6            & 80.7$\pm$1.9          & 79.3$\pm$1.6          & \textbf{74.6}$\pm$1.0 & 75.4$\pm$1.1           & \textbf{80.1}$\pm$2.1    & 71.5$\pm$3.3          & 76.5$\pm$2.0            \\
\textbf{CPL-NoViD} & \textbf{77.9}$\pm$0.4 & \textbf{80.1}$\pm$0.1   & \textbf{80.3}$\pm$1.3   & \textbf{80.7}$\pm$0.4 & \textbf{79.8}$\pm$1.4 & 73.0$\pm$0.6          & \textbf{75.1}$\pm$1.1  & 78.0$\pm$0.6             & \textbf{74.2}$\pm$3.5 & \textbf{80.5}$\pm$2.4   \\ \hline

\multicolumn{11}{c}{\textbf{max context size = 5 (the default setup)}}                                                                                                                                                                                   \\
\textbf{T5-LSTM}                    & 77.8$\pm$1.0                         & \textbf{81.5}$\pm$0.8            & 78.7$\pm$0.8            & 81.8$\pm$1.0      & \textbf{79.3}$\pm$0.8        & \textbf{74.8}$\pm$3.2         & 76.4$\pm$0.9           & \textbf{79.6}$\pm$0.9             & 71.5$\pm$3.6          & 76.8$\pm$1.7            \\
\textbf{CPL-NoViD}                  & \textbf{78.2}$\pm$0.4                         & 80.4$\pm$0.4            & \textbf{80.4}$\pm$0.2            & \textbf{82.2}$\pm$0.4      & 77.3$\pm$1.8        & 72.2$\pm$2.5         & \textbf{78.6}$\pm$0.1           & 78.6$\pm$0.3             & \textbf{77.5}$\pm$1.2          & \textbf{78.9}$\pm$1.2            \\ \hline
\end{tabular}
\addtolength{\tabcolsep}{3.5pt}
\caption{F1 scores (\%) of norm violation detection, with different maximum context sizes, across nine distinct rule types. T5 denotes a variant of T5-LSTM where the LSTM encoder, used for contextual encoding, is removed. The top-performing scores in each category are highlighted in bold. In terms of Macro-F1 scores, CPL-NoViD consistently outperforms T5-LSTM across different maximum context sizes.}
\label{tab:results_abl_context}
\end{table*}

\subsection{Ablation Study I: Evaluating the Addition of Context}

To understand the influence of context on model performance, we conduct an ablation study varying the maximum context sizes. For conversations exceeding the set limit, we truncate to maintain the most recent comments within the context window. For context sizes of one or more, all experimental setups remain consistent with the procedures detailed in the experimental setup.
For scenarios without context, T5-LSTM is adapted by removing the LSTM encoder, with violation predictions based solely on the representation of the concluding comment. This adjustment effectively simplifies T5-LSTM to a basic T5, utilizing only its encoder; CPL-NoViD's context is excluded from the prompt, modifying it to: "In the [subreddit] subreddit, a rule is: [rule]. A comment was posted: [comment]. Does the comment breach the violate rule? [MASK]".

The outcomes of this study are showcased in Table \ref{tab:results_abl_context}.
Our results highlight the pivotal role of context in elevating model performance. Both the baseline and the prompt-based learning models witnessed pronounced improvements with increased context size. Notably, across varied context sizes, CPL-NoViD persistently surpassed T5-LSTM, emphasizing its superior ability to harness contextual nuances. 
This consistent edge that CPL-NoViD maintains illustrates the value of our introduced context-sensitive prompt-based learning model in enhancing norm violation detection.

The substantial improvement exhibited by CPL-NoViD compared to the baseline attests to the benefits of incorporating context as natural language prompts over LSTM-based context encoding. 
These benefits include:
1) \textbf{Long-range dependencies capture:} Pretrained language models are adept at capturing long-range dependencies within text sequences. By including context as natural language in prompts, we enable the model to better understand the relationships between the target comment and its preceding comments. In contrast, LSTMs grapple with the vanishing gradient problem, which impairs their ability to effectively capture long-range dependencies.
2) \textbf{Simplified complexity:} Using natural language prompts to incorporate context eliminates the need for complex context encoders like LSTMs. This simplification reduces the number of model parameters and enhances training efficiency. Conversely, LSTMs have multiple hyperparameters, such as the number of layers and hidden units, which need careful tuning to optimize performance. Effective hyperparameter selection often necessitates manual experimentation and expertise.
3) \textbf{Optimal use of pretrained knowledge:} Pretrained language models, trained on vast amounts of text data, excel in learning syntactic and semantic structures. By incorporating context as natural language in prompts, we leverage the pretrained knowledge of these models to better understand the relationships between the target comment and its context. However, LSTM-based context encoding may not fully exploit this pretrained knowledge and might require additional fine-tuning to perform well on a specific task.
4) \textbf{Enhanced adaptability and robustness:} Our results show that the performance improvement achieved by CPL-NoViD when incorporating context surpasses that of the baseline. This suggests that prompt-based learning with natural language context incorporation exhibits superior adaptability and robustness in handling diverse rule types and community-specific contexts. This makes it ideally suited to the dynamic nature of Reddit communities.

In addition to these benefits, the results in Table \ref{tab:results_abl_context} shed light on the varying influence of context across different rule types. For some rule types, such as ``spam'' and ``off-topic'', context incorporation leads to substantial performance gains, signifying that context is crucial for accurate norm violation detection. Conversely, for rule types like ``incivility'' and ``hate speech'', the impact of context incorporation is less pronounced, indicating that the comment itself may contain sufficient information for detection. These observations underscore the importance of context-aware approaches in effectively managing the diverse spectrum of norms and rules prevalent on platforms like Reddit.

Understanding the role of context in norm violation detection could provide useful insights for designing effective moderation tools. For instance, for rule types where context significantly improves detection accuracy, models could be optimized to focus more on the context, perhaps by allocating more computational resources to context processing or by incorporating more sophisticated context encoding mechanisms. On the other hand, for rule types where the comment itself provides ample information for detection, models could be designed to prioritize the processing of the target comment over the context.

\begin{table*}[ht]
\def\arraystretch{1.2}
\centering
\begin{tabular}{c|ccccc}
\hline
\multirow{2}{*}{\textbf{Method}} & \multicolumn{5}{c}{\textbf{Thread Length}}                                                        \\
                                 & \textbf{2}        & \textbf{3}        & \textbf{4}        & \textbf{5}        & \textbf{6}        \\ \hline
\textbf{T5-LSTM}                 & 81.2$\pm$0.8          & 70.2$\pm$2.0          & 67.1$\pm$1.7          & 63.8$\pm$1.9          & 67.1$\pm$0.8          \\
\textbf{CPL-NoViD}               & \textbf{83.1}$\pm$0.3 & \textbf{73.7}$\pm$0.8 & \textbf{70.6}$\pm$0.5 & \textbf{65.8}$\pm$2.3 & \textbf{74.8}$\pm$0.4 \\ \hline
\end{tabular}
\caption{Micro F1 scores (\%) of norm violation detection on comments in conversation threads of different lengths. For each length, only the conversation threads with the specific length is included, and so the train/val/test data for each different thread length is different. CPL-NoViD outperforms T5-LSTM across different conversation thread lengths.}
\label{tab:results_thread_len}
\end{table*}

\subsection{Ablation Study II: Varying Conversation Thread Length}
Given that CPL-NoViD capitalizes on the surrounding context for norm violation detection, it is imperative to investigate its efficacy across various thread lengths. Specifically, as conversations grow, they tend to evolve in nature, introducing nuances and complexities. An illustrative example is the dynamic role of comments based on their position: despite both being the last comment in their respective threads, a comment appearing in the second place in a conversation thread might be different in nature and context from one appearing in the third place. Analyzing model performance across different thread lengths provides insights into its adaptability and reliability as conversation dynamics change.

For this study, we segment the NORMVIO dataset into subsets based on conversation thread lengths, ranging from two to six\footnote{In NORMVIO there are no threads with a single comment.}. Each subset's train/val/test split is derived from the respective split in the original NORMVIO dataset, retaining only the threads matching the specified length. All other experimental procedures align with those delineated in the experimental setup. The results are tabulated in Table \ref{tab:results_thread_len}. Given the imbalanced distribution of rule types in each subset's test data, where some rules are scarcely represented, we adopt micro-F1 as the evaluation metric.

We observe that both T5-LSTM and CPL-NoViD experience a decline in performance as thread length increases, implying that extracting relevant context from longer conversations poses difficulties for norm violation detection. Notably, CPL-NoViD consistently outperforms T5-LSTM across all thread lengths, emphasizing its enhanced capability in context utilization. Intriguingly, in six-comment threads, both models see a rebound in performance, and CPL-NoViD sees a more significant rebound in its performance for the longest threads. This implies that CPL-NoViD is better at leveraging the extensive context present in these threads, which might include user corrections, repeated sentiments, or other contextual clues that help in identifying norm violations.

\section{Conclusion}
We present a context-aware prompt-based learning method for detecting norm violations during discussions in online communities such as Reddit. Our method named \textbf{C}ontext-aware \textbf{P}rompt-based \textbf{L}earning for \textbf{No}rm \textbf{Vi}olation \textbf{D}etection (CPL-NoViD) outperforms the baseline  in detecting when the discussion violates rules of various types and establishes a new state of the art. By incorporating context through natural language in the prompts, our method mitigates the issues associated with gradient vanishing/exploding and reduces the number of extra parameters required by the learning algorithm.

Our results demonstrate that CPL-NoViD significantly outperforms the baselines in cross-rule-type and cross-community norm violation detection tasks, illustrating the improved generalizability of our model. Moreover, our method shows good performance even in few-shot learning cases, where only a limited number of training examples are available. This is particularly beneficial for smaller communities or less frequent rule types where obtaining a large amount of labeled data can be challenging.

We believe our work has several important implications. First, it highlights the effectiveness of prompt-based learning in the context of norm violation detection and online content moderation. Second, it emphasizes the importance of context-awareness and generalizability in developing models that can adapt to various communities and rules. Lastly, our approach sheds light on the potential of few-shot learning to improve the efficiency of model training and adaptation in real-world scenarios. Together these advances suggest that accurate automated moderation of online discussions is feasible. 

Future research could delve deeper into exploring innovative strategies to further improve the performance and efficiency of the proposed method. These may include the incorporation of external knowledge or the use of more advanced prompting strategies with pre-trained language models. Additionally, it would be worthwhile to investigate how our method can be adapted and extended to other online platforms, broadening its applicability to a diverse range of content moderation tasks. Moreover, advancements in prompt design, such as the use of soft prompts, offer another promising avenue for future research. 

\subsection{Broader Impact and Limitations}
CPL-NoViD prominently showcases its strength in the realms of few-shot learning, cross-rule-type learning, and cross-community learning, where it consistently overshadows T5-LSTM. This underlines CPL-NoViD's agility and ability to generalize in varying data scenarios. Interestingly, in the category of ``hate speech'' detection, T5-LSTM does manifest an advantage over CPL-NoViD. One plausible explanation could be that detecting hate speech relies much less on the context as the semantics in the target comment itself is sufficient for detecting violation of this kind. However, weighing the aggregate performance across all categories, CPL-NoViD exhibits its robustness and superiority. Thus, for applications that demand a comprehensive approach to norm violation detection across a myriad of rule types and intricate conversation contexts, CPL-NoViD, with its context-sensitive prompt-based approach, emerges as the preferred choice.

\section{Acknowledgements}
This material is based upon work supported by the Defense Advanced Research Projects Agency (DARPA) under Agreement No. HR00112290025. Approved for public release; distribution is unlimited.

\section{Release and Access}
Our code and data are publicly available at \url{https://github.com/zihaohe123/norm-violation-detection-prompt-based}.


\bibliography{aaai24}

\section{Ethics Statement}

\subsection{Ethical Considerations}
The proposed CPL-NoViD method for norm violation detection on Reddit aims to improve online moderation and foster healthier online communities by identifying comments that violate platform rules. While the potential positive outcomes of this work are evident, it is crucial to discuss potential negative outcomes and ethical considerations associated with the method's deployment and usage.

One possible negative outcome of our work is the potential for false positives, wherein non-violating comments are mistakenly identified as rule violations. This could lead to unjust actions against users, such as unwarranted bans or removal of content. Conversely, false negatives could allow genuine rule violations to go unnoticed, undermining the goal of creating a safe and healthy community. To mitigate these risks, we encourage the use of human moderators to review and validate the decisions made by our model, particularly in cases with low confidence scores.

Another ethical concern relates to potential biases in the data used for training the model. The public dataset we use, which has already been anonymized, may contain implicit biases reflecting societal prejudices or unequal representation of certain groups. These biases could be inadvertently learned by our model, resulting in discriminatory behavior. To address this concern, it is essential to assess and mitigate the biases in the training data and evaluate the model's fairness across different user groups.

Moreover, the success of our method in detecting norm violations could lead to potential misuse, such as employing the system to enforce censorship or target specific individuals or communities. To prevent such misuse, it is vital to ensure transparency in the deployment of our model and to establish ethical guidelines for its use in real-world applications.

\subsection{Ethics in Data Usage}
Our study involves the use of a public, anonymized Reddit dataset, which includes user-generated content with personally identifiable information (PII) removed. While using an anonymized dataset reduces privacy concerns, it is still essential to consider the ethical implications of using this data in our research.

Regarding the use of the anonymized dataset, it is important to ensure that the data does not contribute to the proliferation of harmful content. We encourage researchers using our model and the dataset to adhere to Reddit's content policy and terms of service, and to use the data responsibly and in compliance with ethical guidelines.

\end{document}